# Plan Development using Local Probabilistic Models

Ella M. Atkins          Edmund H. Durfee          Kang G. Shin

The University of Michigan AI Lab
1101 Beal Ave.
Ann Arbor, Michigan 48109

## Abstract

Approximate models of world state transitions are necessary when building plans for complex systems operating in dynamic environments. External event probabilities can depend on state feature values as well as time spent in that particular state. We assign temporally-dependent probability functions to state transitions. These functions are used to locally compute state probabilities, which are then used to select highly probable goal paths and eliminate improbable states. This probabilistic model has been implemented in the Cooperative Intelligent Real-time Control Architecture (CIRCA), which combines an AI planner with a separate real-time system such that plans are developed, scheduled, and executed with real-time guarantees. We present flight simulation tests that demonstrate how our probabilistic model may improve CIRCA performance.

## 1   INTRODUCTION

To control a complex system, an agent must build and execute a plan that is capable of recognizing state changes due to its own actions or external world events, even when these changes are not completely predictable. The modeling of such systems requires approximations in the form of feature value discretization and/or incorporation of statistical models to describe uncertain phenomena.

Consider an agent capable of safe, fully-automated aircraft flight control from takeoff through landing. To execute a successful flight, the agent must have a set of goals, such as destination airport and intermediate positions, and an accurate model of actions and possible world events. The agent uses this knowledge to build or select plans which dictate control inputs ("actions") as a function of sensor readings ("state feature values"). Because some external events cannot be controlled, plans must include contingencies to handle situations as they arise. For example, since other aircraft share the same skies, onboard radar may detect an aircraft on a collision course at any point during a flight, although the time and location at which traffic will approach is uncertain. An agent must react quickly to avoid this traffic; thus, plans must handle course deviations for collision avoidance. The agent must handle other events including internal systems failure as well as numerous external events such as traffic and significant weather changes. Each event has some chance of occurring as a function of time and state feature values. For example, the probability that traffic will be detected increases with proximity to a busy airport as well as time spent in the air. An agent tries to build and execute plans that yield a high probability of successfully reaching the specified goals. In this paper, we use state probabilities to guide a planner along highly-probable goal paths.

Unfortunately, resource limitations preclude building plans with contingencies for all possible events. For example, the aircraft control agent contains a model of weather phenomena such as precipitation, thunderstorms, wind, etc. High probability events, such as encountering turbulence or rain, should certainly be considered during planning. However, other situations, such as "aircraft flies into a tornado", are possible but very improbable. In fact, just to enumerate all the possible effects of the unlikely event "flying into a tornado" would be difficult — the aircraft could lose any one or more structural components and quickly find itself in any orientation at any altitude. In this paper, we present an algorithm which will remove such low-probability states from consideration during planning, so an ordinary plan will not contain the set of all feature tests and actions to handle "flying into a tornado". We provide a basic detection mechanism for these low-probability states so the agent notices and replans when such a state is encountered.

In this paper, we present a method by which local state probabilities are estimated from action delays and temporally-dependent event probabilities, then used to select highly probable goal paths and remove improbable states. We have implemented these algorithms in the Cooperative Intelligent Real-time Control Architecture (CIRCA). CIRCA combines an AI planner, scheduler, and real-time plan execution module to provide guaranteed performance for controlling complex real-world systems (Musliner, Durfee, and Shin 1995). CIRCA's planner is based on the philosophy that building a plan to handle all world states -- a "universal plan" (Schoppers 1987) -- is unrealistic due to the possibility of exponential planner execution time (Ginsberg 1989), so it uses heuristics to limit state expansion and minimizes its set of selected actions by requiring only one goal path and guaranteeing failure avoidance along all other paths.

We describe CIRCA and compare our work with Markov-based approaches in Section 2, after which we describe probabilistic model specification (Section 3), then computation of local state probabilities (Section 4) and their usage during planning (Section 5). We present results from flight simulation tests (Section 6) that compare plans using our probabilistic model with those developed using no probabilities. We conclude by describing future enhancements to improve our probabilistic model (Section 7).

## 2  BACKGROUND

Many probabilistic planning algorithms have been developed, but most do not consider event probabilities as functions of time. In fact, those of (Kushmerick, Hanks, and Weld 1994) concentrate only on probabilistic properties of actions that may be controlled by the agent, not external events. As described above, events can occur over time without explicit provocation by the agent, and are generally less predictable than state changes due to actions. The Markov Decision Process (MDP) based models (Littman, Dean, and Kaelbling 1995) of (Dean, Kaelbling, Kirman, and Nicholson 1993), (Horvitz and Barry 1995), and (Tash and Russell 1994) may probabilistically model events as "no-op" actions, but do not represent the possibility that individual event probabilities change as a function of time spent in a specific state. This paper explicitly considers temporally-dependent event probabilities, using them to compute state probabilities which help direct planning in CIRCA.

CIRCA (Musliner, Durfee, and Shin 1995) was designed to provide guarantees about system performance with limited sensing, actuating, and processing power. Based on user-specified domain knowledge, CIRCA uses traditional AI techniques to create plans that will keep a system safe (i.e., avoid failure) while working to achieve each plan's goal. CIRCA then uses its knowledge about system limitations to produce a schedule that is guaranteed to execute before deadlines are reached. This scheduled plan is then executed on a separate "real-time" processor.

Figure 1 shows a block diagram of CIRCA. The AI subsystem (AIS) contains both the planner and the scheduler. The "shell" around all AIS operations consists of meta-rules controlling a set of knowledge areas, similar to the PRS architecture (Ingrand and Georgeff 1990). Working memory contains tasks that are ready to be executed. These tasks include planning, downloading plans from the AIS to the real-time subsystem (RTS), and reading/processing feedback data from the RTS.

The CIRCA knowledge base specifies a list of goals which, when achieved in order, will enable the system to successfully reach its final goal. CIRCA executes a planning cycle for each new goal in this list. To minimize domain knowledge complexity, the CIRCA world model is created incrementally based on the initial state set and a group of temporal and action state transitions. Each transition has a name, precondition set, and postcondition set. Action transitions correspond to commands that are explicitly executed by the CIRCA RTS, while temporal transitions correspond to state changes that are not initiated by CIRCA.[1] The planner currently selects actions based on simple criteria, including number of goal features achieved and proximity to failure, and eventually backtracks if a selected action does not ultimately help achieve a goal or avoid failure. CIRCA minimizes its use of memory and time by expanding only states explicitly produced by transitions from initial states or their descendants.

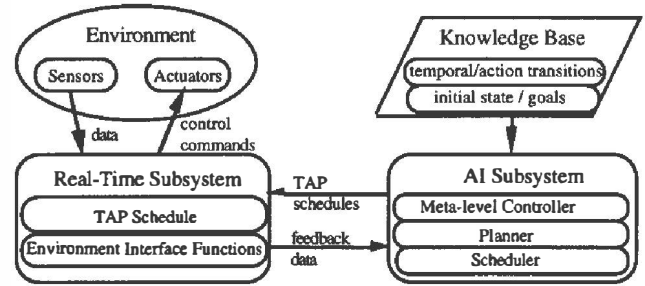

Figure 1: CIRCA Architecture

Figure 2 shows a typical state set expanded during a planning cycle. CIRCA begins planning by selecting one of the initial states and building a list of descendants resulting from temporal transitions (tt). If a tt leads to a state that potentially violates the agent's safe operating envelope, then the state is labelled "failure" and the tt is labelled TTF -- temporal transition to failure. In this case, CIRCA selects an action and associated execution deadline to guarantee avoidance of the TTF. Otherwise, CIRCA may select an action that moves the system closer to the goal. CIRCA continues state expansion for all other initial states and their reachable descendants until at least one goal state is found and all reachable TTFs are guaranteed to be avoided. Note that the planner is minimally satisfied with only one goal path due to tradeoffs between completeness and schedulability (Musliner, Durfee, and Shin 1995). Thus, as shown in the figure, some reachable states (labeled "deadend") do not lead to the goal. These states are "safe" because all TTFs are preempted by actions, but the system has no chance of achieving its goals from those states. Replanning for goal achievement when a deadend state is encountered is discussed in (Atkins, Durfee, and Shin 1996).

CIRCA's control plans are represented as cyclic schedules of test-action pairs (TAPs). Tests involve reading sensors; actions involve sending actuator commands or transferring data between CIRCA modules. When the AIS planner creates a TAP, it stores an associated worst-case execution time and execution deadline to enable safety guarantees. These TAP attributes are then used by a deadline-driven scheduler (Liu and Layland 1973) to create a periodic TAP schedule. If the scheduler is unable to create a schedule that supports all deadlines, the AIS backtracks to the planner, which then selects different actions. This backtrack-(select actions) cycle repeats until either the scheduler succeeds or until the planner can find no actions that avoid failure and reach the goal. We propose using state transition probabilities to assist the scheduler by removing the lowest probability states when backtracking, thus allowing removal of all guaranteed actions planned only to preempt failure from highly improbable states. If one of the low-probability "removed" states is actually reached, CIRCA detects it and replans as discussed in (Atkins, Durfee, and Shin 1996).

---

[1] Previously (Musliner, Durfee, and Shin 1995), CIRCA contained three transition types: action, temporal, and event. "Events" can occur instantaneously while "temporals" have a non-zero delay. We now model events and temporals as "temporal transitions", with differences specified using probability functions as described in Section 3.

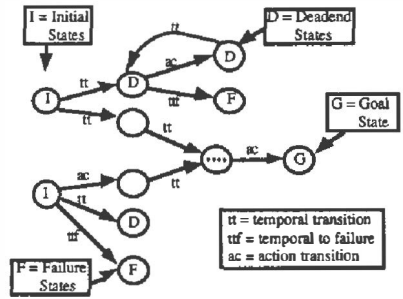

Figure 2: State Diagram Expanded during Planning

As discussed above, CIRCA separates the planning phase from plan execution to permit the building of a relatively complete plan while providing action execution time guarantees.[2] In MDP-based models, plan development and execution are often not separated, so methods to restrict planning time must be used before meeting hard execution deadlines. For example, (Dean, Kaelbling, Kirman, and Nicholson 1993) build an "envelope" of states based on initial state and selected actions, but this envelope may model an inadequate state set with a short plan completion deadline. Alternatively, (Boutilier and Dearden 1994) discuss feature abstraction to reduce state-space size, but significant speedup occurs only if abstraction is extensive. MDP utility allows a planner to trade off execution costs with the benefit of continued planning. Such utilities may be constructed to promote planner termination before reaching a deadline (as in (Horvitz and Barry 1995)), but plan completeness may suffer if the cost of computation time is given sufficient importance to meet hard deadlines.

## 3 USER SPECIFICATION OF TRANSITION PROBABILITY

The CIRCA planner builds the reachable state set based on action and temporal transitions specified in the domain knowledge base. Figure 2 shows that any state may have multiple outgoing transitions. We wish to calculate the probability of reaching a state from the probability of its parent state(s) so that we can propagate state probabilities from initial states throughout the network. This can be done if we can explicitly describe the cumulative probability of each state transition as a function of time.

---

[2] Separation of planning and plan execution was used to allow unrestricted planning time while meeting execution deadlines. Our recent work with handling unexpected states (Atkins, Durfee, and Shin 1996) will require us to restrict planning time, using methods similar to those described in MDP work, but CIRCA's planner will still require few constraints so long as states remain within the "handled" region.

We use a simple model for action transition probabilities. As with the previous version of CIRCA (Musliner, Durfee, and Shin 1995), we assume action transitions will affect state features following a constant delay after being executed on the RTS. Figure 3 shows the cumulative probability function used for actions as a function of time.[3] To specify this function, the user computes a delay ($t_{delay}$) between the time the action is initiated (time=0) and the time at which the action changes state features (time=$t_{delay}$). Then, the total delay between reaching the state prompting action execution and the time that action affects state features is:  **t (total delay) = $t_{delay}$ +** (delay between when the state is reached and when this action begins executing in the schedule). In Section 4, we approximate the second term in this equation using calculated deadlines and average schedule execution times.

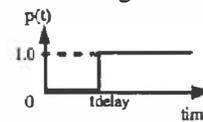

Figure 3: Action Transition Cumulative Probability

Because they are not explicitly commanded, we cannot assume such tight control over temporal transitions, some of which may not be precisely modeled. For this reason, we allow the user to define a cumulative probability function for each temporal transition, where time=0 is defined as the time at which that transition's preconditions were first satisfied. Figure 4 shows two examples of temporal transition probability functions and their associated cumulative probability functions. In Figure 4a, the transition has a high probability of occurring immediately. This probability decays over time, leaving a cumulative probability asymptote of **Pmax** < 1.0. The value (1 - **Pmax**) corresponds to the probability that this event will never occur. As an example, consider the state in which an aircraft collision avoidance alarm (indicating nearby traffic) has just sounded. The probability $p(t)$ that the transition "collide with other aircraft" will occur immediately jumps to its maximum value, but decays in time, since either the other aircraft will pass or else the collision will have already happened.

Figure 4b shows an example for which a delay occurs between the time the state is reached and the time this transition may happen (i.e., $p(t) > 0$). The asymptote of the cumulative probability function is 1.0, indicating this transition will occur if given sufficient time. A simple example of this transition type is travel between distinct locations. Suppose an aircraft flies along the coast from Los Angeles to Portland, Oregon. At a point along the flight, the aircraft state changes to "Location: San Francisco", at which time the aircraft heads directly for Portland. The probability that the temporal transition "Arrive in Portland" will occur is near zero for a certain amount of time, even with a tremendous tailwind and maximum thrust. The peak value of $p(t)$ occurs at the expected arrival time based on average calculations, while the width of p(t) increases as the uncertainty in wind, aircraft performance, and/or course deviations increases.

---

[3] We assume all state features are observable and that if an action is initiated, it will be executed properly. Otherwise, we could not specify action probabilities that reach 1.0.

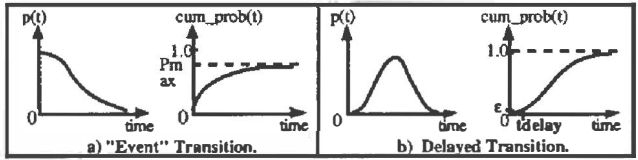

Figure 4: Temporal Transition Probability Functions

Because we allow multiple temporal transitions from any state, probabilistic dependencies between these transitions must be considered. These dependencies exist because the occurrence of one temporal transition changes the current state, thus no other temporal transition may later occur from that state. In previous CIRCA versions, the number of temporal transitions modeled in the knowledge base was minimized by making preconditions minimal so that temporal transitions could occur in different combinations from many states. However, we now must accurately capture transition probability dependencies, so the user must make preconditions more specific, increasing the number of temporal transitions in the CIRCA knowledge base. The following procedure defines how the user specifies temporal transitions and their probabilities:

(1) Define temporal transition "sets", each of which contains all temporal transitions with a specific set of preconditions. Each set's preconditions must be sufficiently specific such that no state can match the preconditions of two different transition sets.
(2) For each set, specify the probability function for each transition. The sum of all cumulative probability function asymptotes in each set must be ≤ 1.0 (100%). We assume the user has sufficiently restricted preconditions to explicitly define any features on which transition probabilities depend.

The above procedure may become complicated when many different sets of temporal transitions exist. To-date, our tests have used relatively simple models with few features (see Section 6); otherwise, we would probably still be building our knowledge base. Unfortunately, such complexity is present in many probabilstic models, including ours and MDPs (Littman, Dean, and Kaelbling 1995), in which at least a constant probability must be specified for transitioning from each state to any other state. We are working to automate probability dependency calculation, building on methods such as (Haddawy 1994).

One of CIRCA's main premises is that of guaranteed failure avoidance. In previous versions, guarantees required that, for any state with a temporal transition to failure (TTF), an action must be scheduled such that its t (total delay) is less than the delay of a corresponding TTF (i.e., the delay before cum_prob ε is reached as shown in Figure 4b). With this restriction, the action preempts the TTF making the TTF's probability near zero with small ε. The algorithms described in Sections 4 and 5 enable us to compute state probabilities not just by considering constant probability transitions between the states, but also by considering when any temporal and action transitions will happen in relation to each other. We use these state probabilities to allow the removal of highly improbable states (i.e., "airplane flies into a tornado") and selection of highly probable goal paths.

## 4 LOCAL COMPUTATION OF STATE PROBABILITY

State probabilities are computed recursively during state expansion, with a "parent" state and applicable outgoing transitions used to determine "offspring" probabilities. The planner begins with an initial state set and no knowledge of relative probabilities within this set, so we assume a non-informative prior distribution. The planner expands each initial state, using the available transitions to develop the set of offspring states and initialize their probabilities. Offspring states eventually become parent states to be expanded, continuing until all reachable states (i.e., states with probability > ε) have been expanded.

A set of zero or more action and temporal transitions match the preconditions of any parent state. The states resulting from all matching temporal transitions and any planner-selected action are the offspring set. Figure 5 illustrates the three possible situations. In Figure 5a, a TTF exists, so the planner has chosen a preemptive (guaranteed) action. Offspring states, P1-Pn, result from that action and all applicable temporal transitions. State Pn is a failure state that must have probability less than a small value ε. Figure 5b illustrates the case when a non-preemptive action is selected. Offspring states result from that action and the (n-1) temporal transitions. Finally, Figure 5c illustrates the case when no action is selected, a possibility if no TTF exists and the planner selects no action along a goal path. In this case, all n offspring states result from temporal transitions.

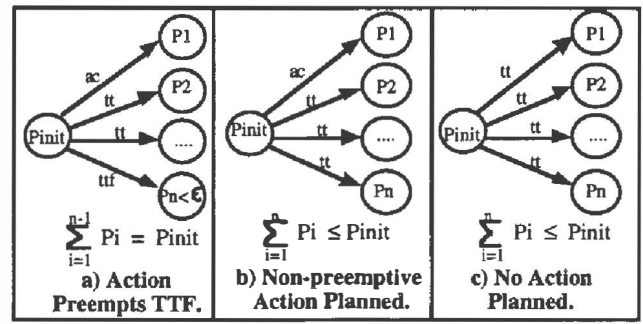

Figure 5: Possible Transitions from Parent to Offspring

The algorithm in Table 1 is used to locally compute probabilities for each reachable state. First, the planner selects the most probable state to be expanded, then selects an appropriate action (if any). Next, the planner builds a list of possible offspring, resulting from temporal transitions with preconditions matching the parent state plus the action if one has been chosen. Transition probabilities are functions of time, but we wish to specify state probabilities as constant values because they are used for ranking the states during planning (see Section 5). To determine state probability values, we select a time at which each transition's cumulative probability critically affects the state. For complete accuracy, we would use the schedule to predict average action execution times. However, the plan must be developed before scheduling, and we need the state probabilities from the schedule to develop the plan. We thus make heuristic approximations to select a critical time (**t**) based on action execution deadlines and timings for feature tests and actions.

Selection of **t** depends on which of the three Figure 5 cases is present. In case (a), there is a TTF so we require a guaranteed action. With a fully-populated schedule, the action will execute with near 100% probability just before the TTF has probability ε of occurring, so we set this as critical time **t**. In case (b), the action is not preemptive, so there is no associated execution deadline. Because this action is executed only when extra time is available, we would need to know the schedule and progression of world states to accurately predict execution frequency. The planner knows neither of these quantities, so we assume execution time delay (**t**) for an "average" case (i.e., half of all actions and feature tests are executed; half of the schedule has executed). Mathematically, **t = a\*n/8 + b/4 + tdelay**, where **tdelay** = (action execution delay), **a** = Σ (feature test execution times), **n** = (number of actions available), and **b** = Σ (action execution times). We selected **t** to average two approximation errors: **t** will be too high because it is unlikely that half of all possible actions will be executed in a single schedule cycle (i.e., a plane cannot climb and descend at the same time), but **t** is low because, with a full schedule of guaranteed TAPs, several schedule cycles may pass before this non-preemptive TAP has time to execute. Finally, in case c, no action has been scheduled -- the state will remain unchanged until some temporal transition occurs, so we consider the asymptotic values of the transition probability functions (i.e., t->infinity). Using the critical time **t** for each case, the planner then creates a list of cumulative offspring probabilities. For each offspring resulting from a temporal transition, cumulative probability is the value of that temporal transition's probability function at time **t**. For an offspring resulting from an action, let **p** = (1 - Σ (all offspring probabilities resulting from temporal transitions)). Then, with a preemptive action, let **Pi = p** (because we are certain the action will execute by **t**, the execution deadline), and for a non-preemptive action, let **Pi = p/2** (because we assume average case, and there is no guarantee a nonpreemptive action will ever execute). All offspring probabilities are then scaled to reflect the parent's probability (**Pinit**).

Figure 6 shows the three possible offspring state cases with respect to pre-existing parent states: 6a) offspring state is new, 6b) offspring state exists, has some nonzero probability, but has not been expanded, and 6c) offspring state exists and already has a probability ≥ the current parent state, thus has already been expanded. In case 6a), the probability due to new parent **P1** is the only contribution to offspring probability, so **Piold=0** (Table 1, step 6) and **Pi** (Table 1, step 5) is the total offspring probability. In case 6b), some other parent(s) exists, but the offspring has not yet become a parent, so its probability can be increased without propagating the new value through other offspring states. For this case, we simply sum the old (**Piold**) and new **Pi** contributions to compute offspring probability (Table 1, step 6).

In Figure 6c, the new probability addition (**P1**) affects a state that has already been expanded (**P2**). Modifying **P2** would require updating **P2**'s immediate and downstream offspring that have been generated. In fact, **P1** could be a descendant of **P2**, in which case a cycle would exist. We currently do not propagate probabilities in this case -- such propagation would require the detection and handling of directed cycles, a common occurrence in many plans (i.e., aircraft flies in a holding pattern before landing). Since probabilities are always positive, this error only results in underestimation (as shown by the example in Section 6), thus probabilities are never falsely inflated.

The two major approximations in the local probability computation algorithm -- estimating constant state probability values at "critical" times, and propagating offspring probability only when the offspring has not yet been expanded -- allow state probabilities to be computed quickly and accurately on average. Future improvements for both approximations are discussed in Section 7.

Table 1: State Probability Calculation Algorithm

| |
|---|
| 1. Select the most probable state for expansion and let **Pinit** be this state's probability. (**O(m)**) |
| 2. Select an action by scoring all potential action candidates. (**O(nf\*na)**) |
| 3. Create a list of offspring states for temporal and action transitions. (**O(nt)**) |
| 4. Compute critical time (**t**) for transition probabilities. Case a): Preemptive action execution deadline; Case b): Non-preemptive action average delay time; Case c): Temporal transition asymptotic probability. (**O(1)**) |
| 5. Create a list of cumulative probabilities for the offspring states (**O(nt)**). |
| 6. Scale each probability by **Pinit** (**Pi = Pi \* Pinit**). (**O(nt)**) |
| 7. For each unexpanded offspring state, add any previously existing probability due to other parent states to the new value (**Pi = Pi + Piold**). (**O(nt)**) |
| Overall complexity: (**O(nt)**) **O(m + nf\*na + nt)**, where **m**=number of unexpanded, reachable states that could become parent states, **nf**=number of features, **na**=number of action transitions, **nt** = number of temporal transitions) |

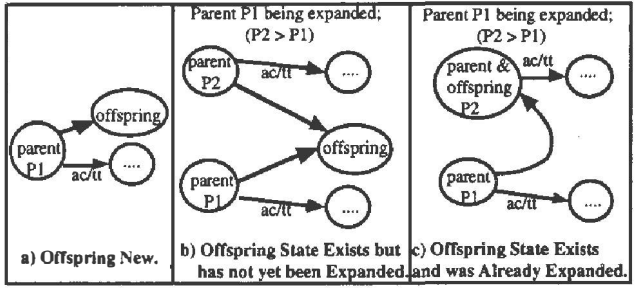

Figure 6: Offspring State Connectivity Cases

# 5 PLAN CREATION

We use state probabilities in two ways: finding highly-probable goal paths and removing improbable states. In this section, we consider three classes of states: goal-reaching, deadend (as described in Section 2), and removed (states that are reachable but removed during planning). Figure 7 shows plots of probability range for different state class types, where Pmax is the maximum probability of reaching any one state. Figure 7a depicts what we consider the "ideal" achievement from using state probabilities. In this case, the planner searches for goal paths until all states above a probability threshold P2 can reach the goal, and states below a much lower threshold P1 are removed. With sufficient resources, P1 and P2 may be set to zero and all states may reach the goal. As resource availability becomes more restrictive, we envision a planner that can use scheduler feedback to iteratively increment P1 and P2 to achieve this "ideal" diagram, which is theoretically possible so long as (1) the knowledge base is complete and correct and (2) the probability contribution of all removed states (i.e., p<P1) to any single offspring is not sufficient to push that state's probability above either the P1 or P2 probability threshold. It is virtually impossible to guarantee these two conditions, so in reality, the "ideal" case can only be approximated. Also, for these diagrams, we assume some action sequence will allow the system to reach a goal from any state, so the planner can choose whether a state is deadend or not. In reality, there may be some states that are deadend by necessity, and those states may fall above the P2 threshold, even in the "ideal" case.

Figure 7b shows the state class probability plot we approximately achieve in the new version of CIRCA. The planner attempts to find the single most-probable goal path by performing a best-first search in decreasing order of state probability, but because we do not have perfect state probabilities and high probability states may not all lie on the same path, the highest probability states include both goal-reaching and deadend states. The threshold P2 is determined by the lowest probability state on the single goal path. P1 is initially set by the user then incremented automatically as necessitated by scheduling constraints (i.e., the planner can't schedule all reactions associated with "flight into a tornado").

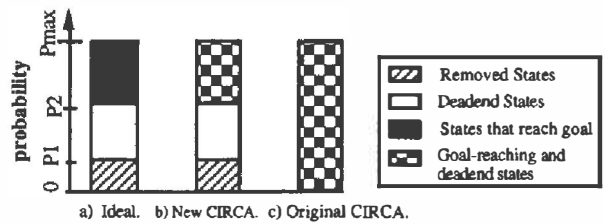

Figure 7: Probabilities for the Different State Classes

In previous versions of CIRCA (Musliner, Durfee, and Shin 1995), the planner expanded states in depth-first order. The planner selected actions primarily to avoid TTFs and secondarily to achieve goals. Probabilities were not considered, so no states were removed and any state could be either goal-reaching or deadend, as illustrated in Figure 7c. In the worst case, the only goal-reaching states would have probability near 0, thus deadend states would almost certainly be reached. Although the "new" CIRCA (Figure 7b) doesn't perform "ideally" (Figure 7a), it has a better chance of reaching its goals than the "old" CIRCA because probability considerations prevent the worst case. Quantifying how much better the new CIRCA performs involves evaluating the probabilistic model for a given domain, as well as estimating the presence of cycles, etc., that degrade calculated state probability accuracy.

Figure 8 shows state diagrams in which a low-probability transition leads out of the initial state. For clarity, the reader may think of the low-probability **tt** as "flight into a tornado", and downstream actions necessary to avoid a subsequent crash. In the first planner iteration (Figure 8a), all states are kept, the goal state is found along the higher-probability path, and actions (e.g., climb) are planned to avoid failure (state "F"). When the planner determines it cannot schedule all preemptive actions, P1 is incremented such that all states downstream of the low-probability temporal transition are removed (Figure 8b) (i.e., the states after "flight into a tornado"). Without state removal, CIRCA's planner may completely fail since it requires plans with guaranteed failure avoidance. This problem was recognized and discussed in (Musliner, Durfee, and Shin 1995), but no formal algorithms for considering characteristics such as probability were presented. We give the planner a better chance to reach its goals by reducing the number of required actions via low-probability state removal. In the unlikely case that a removed state is reached, we (Atkins, Durfee, and Shin 1996) have developed an algorithm to detect and replan so that such an occurrence does not spell certain doom.

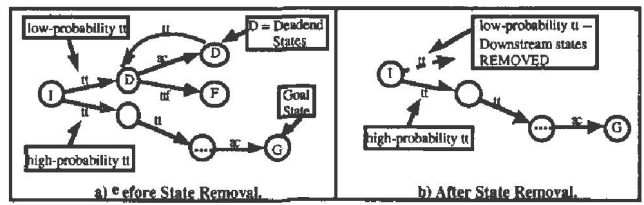

Figure 8: "Removed state" Illustration

## 6  Tests from Flight Simulation

The ACM aircraft flight simulator (Rainey 1994) was chosen to test the new CIRCA probabilistic model. We selected the "flight around a pattern" task for our automated aircraft, as illustrated in Figure 9. A series of subgoals to arrive at different "fixes" allow the planner to consider a small flight segment in each planning cycle. The aircraft CIRCA knowledge base contains several high-level features necessary for our task, including altitude (ALT), location (LOC; "fix" in Figure 9) last reached, heading (HEAD), gear status (GEAR), and the status of other traffic on final approach. Action transitions are specified to control navigation settings (OBS – omnibearing selector), altitude, and heading. A low-level control system interfaces the high-level CIRCA actions with simulator actuators, as discussed in (Atkins, Durfee, and Shin 1996). Temporal transitions model flight to new locations, altitudes, traffic, and gear status changes.

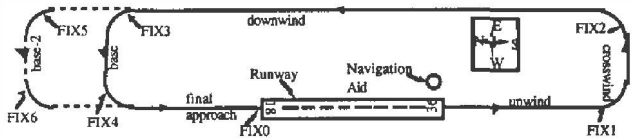

Figure 9: Flight Pattern Flown during Simulation

We concentrate on the "Fly from FIX3 to FIX4" subgoal for this discussion; other flight segments are handled similarly. Figure 10 shows the state diagram resulting from planning with probabilities. State probabilities are represented above each state. Action transitions are represented by bold arrows, while plain arrows represent temporal transitions. As shown, the goal path involves turning left to a heading (HEAD) of West (W), setting the OBS to the next fix, then flying until reaching FIX4. Any loss of altitude that might eventually result in a crash is countered by a guaranteed "climb" action. The improbable event that traffic might be present on final approach would result in a deadend state. As discussed in Section 4, inaccuracies in probability result from estimating schedule execution time and not updating probabilities with cycles such as the "loss-of-altitude" / "climb" loops. In this case, the schedule actually has slack time and cycles lead back to the goal path, so CIRCA is underestimating goal path probabilities.

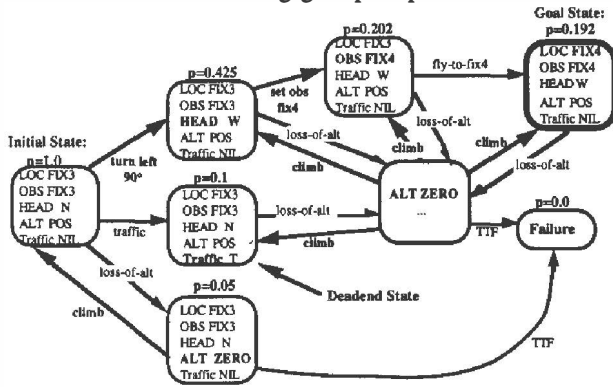

Figure 10: "Fly FIX3 to FIX4" States using Probabilities

We compared the Figure 10 result to plans created when using the "old" CIRCA with no model of probability. Two goal paths were possible -- the one shown in Figure 10, and the longer, less probable path shown in Figure 11. In the Figure 11 path, the aircraft continues from FIX3 to FIX5, FIX6, then to FIX4. The Figure 11 probabilities were calculated using the algorithm from the "new" CIRCA, so they are less than actual values, but not by as much because the schedule contains less slack time. By randomly reordering the states to be expanded (corresponding with starting the "old" CIRCA's depth-first-search along different branches), we were able to create each of the two plans shown here. One cannot easily predict which plan the "old" CIRCA will produce.

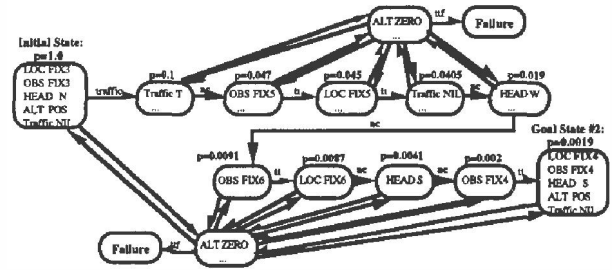

Figure 11: "Fly FIX3 to FIX4" States -- No Probabilities

To illustrate the utility of removing states, we incorporated a feature "gear" with values "up" and "down", corresponding to the position of the aircraft landing gear. All initial states had value "down", and the planner chose no action to raise the gear. However, it is highly improbable but possible that the extended landing gear could fail anytime during flight, thus a low-probability "gear-up" transition could be applied to any state. The result of applying the "gear-up" transition to all states is shown in Figure 12 -- the state diagram effectively doubles in size, even though half the states have very low probability. By specifying a cutoff probability (P1 from Section 5) greater than the asymptote of the "gear-up" probability function, the "new" CIRCA expanded half (or less) of the states that the "old" CIRCA had to expand.

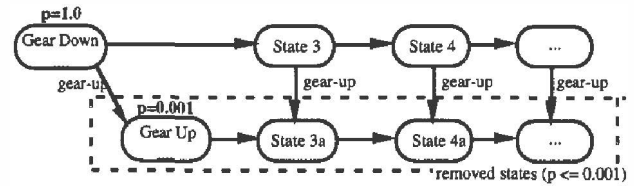

Figure 12: Illustration of Improbable State Removal

Figure 13 summarizes the "Fly FIX3 to FIX4" plans produced by the "new" CIRCA, the "old" CIRCA following the Figure 10 goal path, and the "old" CIRCA following the Figure 11 goal path. The big wins for the new CIRCA are in number of states expanded, due to both improbable state removal and shorter average goal path, and in average probability that the goal will be reached, since the user does not know which goal path will be chosen in the "old" CIRCA.

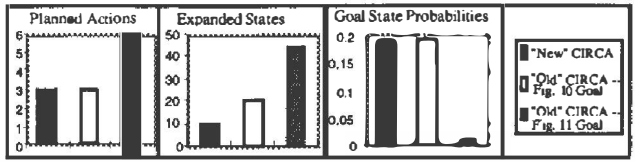

Figure 13: "Fly FIX3 to FIX4" Plan Results

# 7 SUMMARY AND FUTURE WORK

This paper describes a general temporally-dependent probabilistic model for plan development. This model was implemented and tested in CIRCA. The user specifies event probabilities and action delays in the domain knowledge base, then CIRCA locally computes offspring state probabilities based on any selected action and applicable temporal transition probability functions. These state probabilities are used to select highly-probable goal paths and to remove improbable states from consideration when planning or scheduling difficulties arise.[4] Our approach differs from traditional MDP-based planners because our transition probabilities are functions of time and we accommodate hard action execution deadlines by explicitly scheduling each plan.

We have tested these algorithms by assessing performance differences between the "old" (no probabilistic model) and "new" CIRCA during the "flight around the pattern" task. In our preliminary tests, by removing the low-probability "gear up" states and directing search along a highly probable path, the "new" CIRCA reliably develops a plan which achieves the goal with relatively high probability and expands fewer states than the "old" CIRCA.

While these results are promising, we continue to improve our probabilistic model. We may better model action transition delays using time functions analogous to the temporal transition probability functions, similar to action probability models proposed in (Haddawy 1994). We are working to incorporate such functions into our state probability calculation algorithm. Next, we need to better quantify state probability accuracy, which is difficult to estimate during planning. After planning is complete, however, we can compare actual schedule execution parameters (i.e., number of planned actions, test features, schedule cycle time) with estimated averages to dynamically evaluate accuracy of the critical times assumed during state probability computation. Another source of inaccuracy is the absence of probability propagation through previously-expanded states, because we must consider the possibility of encountering directed cycles.[5] We hope to incorporate approximate methods that can detect and handle cycles of a limited size.

---

[4] We have chosen states to remove based solely on probability in this paper. In future work, we hope to look at a more general utility function for selection of states to remove and expand next. Such a function might include proximity to failure and minimum time before being reached (i.e., time horizon) as well as probability.

[5] CIRCA cannot produce undirected cycles because of time considerations -- once a state is reached, the past cannot be reencountered unless a new set of transitions explicitly leads back, producing a directed cycle.

## Acknowledgments

This work was supported under NSF grant IRI-9209031.

## References

E. M. Atkins, E. H. Durfee, and K. G. Shin, "Expecting the Unexpected: Detecting and Reacting to Unplanned-for World States," to appear in *Proceedings of AAAI Workshop on Theories of Action and Planning: Bridging the Gap*, August 1996.

C. Boutilier and R. Dearden, "Using Abstractions for Decision-Theoretic Planning with Time Constraints," *Proceedings of AAAI*, pp. 1016-1022, 1994.

T. Dean, L. P. Kaelbling, J. Kirman, and A. Nicholson, "Planning with Deadlines in Stochastic Domains," *Proceedings of AAAI*, pp. 574-579, July 1993.

M. L. Ginsberg, "Universal Planning: An (Almost) Universally Bad Idea," *AI Magazine*, vol. 10, no. 4, Winter 1989.

P. Haddawy, *Representing Plans Under Uncertainty: A Logic of Time, Chance, and Action*, Springer-Verlag, Berlin, 1994.

E. Horvitz and M. Barry, "Display of Information for Time-Critical Decision Making," in *Proceedings of UAI-95*, August 1995.

F. F. Ingrand and M. P. Georgeff, "Managing Deliberation and Reasoning in Real-Time AI Systems," in *Proc. Workshop on Innovative Approaches to Planning, Scheduling and Control*, pp. 284-291, November 1990.

N. K. Kushmerick, S. Hanks, D. Weld, "An Algorithm for Probabilistic Least-Commitment Planning," *Proc. of AAAI*, pp. 1073-1078, July 1994.

M. L. Littman, T. L. Dean, and L. P. Kaelbling, "On the Complexity of Solving Markov Decision Problems," *Proceedings of UAI-95*, August 1995.

C. L. Liu and J. W. Layland, "Scheduling Algorithms for Multiprogramming in a Hard Real-Time Environment," *Journal of the ACM*, vol. 20, no. 1, pp. 46-61, January 1973.

D. J. Musliner, E. H. Durfee, and K. G. Shin, "World Modeling for the Dynamic Construction of Real-Time Control Plans", *Artificial Intelligence*, vol. 74, pp. 83-127, 1995.

R. Rainey, *ACM: The Aerial Combat Simulation for X11*. February 1994.

M. J. Schoppers, "Universal Plans for Reactive Robots in Unpredictable Environments," in *Proc. Int'l Joint Conf. on Artificial Intelligence*, pp. 1039-1046, 1987.

J. Tash and S. Russell, "Control Strategies for a Stochastic Planner," *Proc. of AAAI*, vol. 2, pp. 1079-1085, 1994.